\documentclass{article}

\usepackage{arxiv}

\usepackage[utf8]{inputenc} % allow utf-8 input
\usepackage[T1]{fontenc}    % use 8-bit T1 fonts
\usepackage{hyperref}       % hyperlinks
\usepackage{url}            % simple URL typesetting
\usepackage{booktabs}       % professional-quality tables
\usepackage{amsfonts}       % blackboard math symbols
\usepackage{amsmath}       % advanced math commands
\usepackage{nicefrac}       % compact symbols for 1/2, etc.
\usepackage{microtype}      % microtypography
\usepackage{lipsum}
\usepackage{graphicx}
\renewcommand{\textit}[1]{#1}
\renewcommand{\emph}[1]{#1}
\graphicspath{ {./images/} }

\title{OSCToM: RL-Guided Adversarial Generation for High-Order Theory of Mind}

\author{
 Sharmin Sultana Srishty \\
  Department of Computer Science\\
  BRAC University\\
  \texttt{sharmin.sultana.srishty@g.bracu.ac.bd} \\
   \And
 Kazi Mahathir Rahman \\
  Department of Computer Science\\
  BRAC University\\
  \texttt{kazi.mahathir.rahman@g.bracu.ac.bd} \\
  \And
 Malaika Parizat Sakkhi \\
  Department of Computer Science\\
  BRAC University\\
  \texttt{malaika.parizat.sakkhi@g.bracu.ac.bd} \\
  \And
 Samia Shahid Prianna \\
  Department of Computer Science\\
  BRAC University\\
  \texttt{samia.shahid.prianna@g.bracu.ac.bd} \\
  \And
 Shaikhul Islam Sinat \\
  Department of Computer Science\\
  BRAC University\\
  \texttt{shaikhul.islam.sinat@g.bracu.ac.bd} \\
}

\begin{document}
\maketitle

\begin{abstract}
Large Language Models (LLMs) perform well on many language tasks, but their Theory of Mind (ToM) reasoning is still uneven in complex social settings. Existing benchmarks, including \textit{ExploreToM}, do not always test the recursive beliefs and information asymmetries that make these settings difficult. This paper presents \textbf{OSCToM} (\textbf{Observer-Self Conflict Theory of Mind}), an approach for modeling nested belief conflicts in LLM-based ToM tasks. The key case is one in which an observer's view of another agent conflicts with the observer's own belief state. Such cases go beyond simple perspective-taking and require recursive, multi-layered reasoning. OSCToM combines reinforcement learning (RL), an extended domain-specific language, and compositional surrogate models to generate observer-self conflicts. In our experiments, OSCToM-8B gives the best overall result among the systems tested. It improves on the reported \textit{ExploreToM} results on FANToM and remains competitive on Hi-ToM and BigToM. On the information-asymmetric \textbf{FANToM} benchmark, OSCToM reaches \textbf{76\%} accuracy, compared with the \textbf{0.2\%} reported by \textit{ExploreToM}. The data-synthesis procedure is also \textbf{6x} more efficient, indicating that targeted training data can help smaller models handle advanced cognitive reasoning. The project code is available at \url{https://github.com/sharminsrishty/osct}.
\end{abstract}


\section{Introduction}

Theory of Mind (ToM) is the ability to reason about the beliefs, intentions, and knowledge of other agents. The term was introduced in studies of primate social intelligence \cite{premack1978mind} and later became central to work on human social cognition. ToM supports many forms of social interaction, including cooperation, persuasion, and deception. In everyday reasoning, people do not only track facts; they also track what others know, what others falsely believe, and how those beliefs differ from reality. For Large Language Models (LLMs), this ability is an important part of social reasoning. It also shifts evaluation beyond fluent text generation toward the question of whether a model can represent and update mental states.

\begin{figure}[h]
    \centering
    \includegraphics[width=0.85\linewidth]{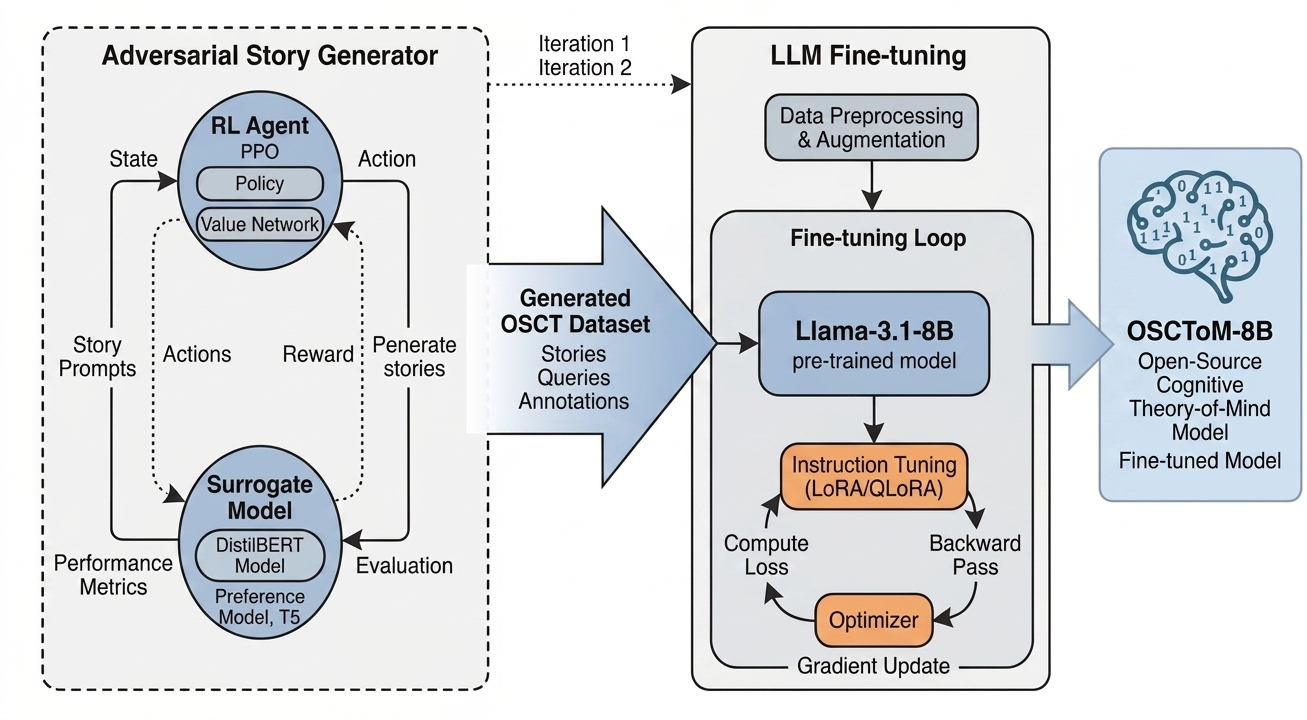}
    \caption{The OSCToM system architecture. The left module depicts the Adversarial 
    Story Generator (combining RL and surrogate evaluation) which constructs the OSCT 
    dataset. The right module depicts the subsequent LLM Fine-Tuning pipeline, 
    resulting in the final OSCToM-8B model.}
    \label{fig:architecture}
\end{figure}

Early NLP work tested ToM mostly with static, hand-written vignettes. A common example is ToMi \cite{le2019tomi}, which evaluates simple false-belief tracking. These tasks were useful, but they are now less reliable as model scale and training data have increased. A model may score well by relying on familiar story patterns or surface cues rather than by tracking beliefs in a consistent way \cite{mindreading2023kosinski, sap2023review, binz2023using}. This creates a reasoning gap: models can perform well on standard examples but fail when the same logic is presented through a slightly different narrative structure.

Programmatic and adversarial generation offers one way to reduce these limits. The \textit{ExploreToM} framework \cite{sclar2024exploretom}, for example, uses a domain-specific language (DSL) and A* search to synthesize stories under difficult information conditions. This is an important step, but search-based generation still has limitations. A* search is constrained by the predefined search space and does not adapt its policy from experience. It may also reward informational volume more than the specific structure of the belief conflict. As a result, models can still fail on narratives that require deep information asymmetry and multi-layered recursive tracking, such as cases where one agent reasons about another agent's belief about a third agent's intention.

We focus on a type of reasoning that we call \textbf{Observer-Self Conflict}. This state appears when an observer attributes a belief to another agent while holding a different belief internally. The conflict is not only between a character and the true world state, but also between nested perspectives inside the observer's model of the situation. Such cases are common in complex social reasoning, but they are difficult to generate and verify at scale. OSCToM addresses this problem by adding these conflict structures to an optimized LLM generation and training pipeline.

Instead of relying only on fixed search heuristics, OSCToM treats story generation as an optimization problem. A DQN-based reinforcement learning generator learns to move through an extended DSL that supports 4th-order belief tracking and deceptive mental states. Because direct LLM verification is expensive for this type of generation, we use a \textbf{compositional surrogate pipeline}. The pipeline contains six specialized modules that estimate factual and belief difficulty with much lower cost than full LLM-based verification. This surrogate-guided design makes large-scale adversarial data synthesis practical and gives a \textbf{6x} overall efficiency gain in pipeline execution.

The resulting model, \textit{OSCToM-8B}, is trained with a two-stage curriculum. In our experiments, it shows strong 4th-order reasoning performance and outperforms much larger models in several settings. On the information-asymmetric \textbf{FANToM} benchmark \cite{zheng2023fantom}, OSCToM-8B reaches \textbf{76\%} accuracy, compared with the \textbf{0.2\%} reported for the \textit{ExploreToM} baseline.

The rest of this paper is organized as follows: Section 2 reviews related work in Theory of Mind and adversarial benchmarks; Section 3 details the OSCToM framework, the Extended DSL, and our RL-guided generation policy; Section 4 presents our experimental setup and results across global benchmarks; and Section 5 discusses the implications of our findings for the future of social AI.

\section{Related Work}

Theory of Mind (ToM) is usually defined as the ability to attribute mental states, such as beliefs, intentions, and knowledge, to oneself and to others. It was first formalized in primate psychology \cite{premack1978mind} and later became an important marker in studies of human development \cite{wimmer1983beliefs, baroncohen1985mind}. In the domain of Artificial Intelligence, evaluating whether computational systems possess this capability has moved from philosophical inquiry into an empirical benchmark of social intelligence. Early neural evaluations of ToM focused on simple story-comprehension tasks, suggesting that social intelligence could emerge as a side-effect of large-scale language modeling \cite{bubeck2023sparks}. These initial benchmarks mainly used the \textbf{ToMi} dataset \cite{le2019tomi}, which parameterized the classic Sally-Anne false-belief test into simple, linear text vignettes. While these early evaluations suggested rising competence in models like GPT-3, later studies showed that such performance was fragile. Researchers showed that models often relied on shallow heuristics, spurious correlations, and narrative pattern matching rather than on a stable internal model of causal belief states \cite{mindreading2023kosinski}. In particular, targeted behavioral studies found that small textual changes, which would be easy for a human reader, could cause large performance drops in state-of-the-art models \cite{ullman2023large}. This discrepancy between perceived fluency and actual cognitive modeling led to a broader critique in the field \cite{sap2023review, binz2023using}, highlighting a persistent "reasoning gap" that required a shift toward more complex, multi-layered reasoning benchmarks designed to test these models more effectively.

These findings motivated harder benchmarks for higher-order and recursive belief reasoning. Studies such as \textbf{Hi-ToM} \cite{zhu2023hitom} and \textbf{BigToM} \cite{gandhi2023bigtom} aimed to test second- and third-order recursive beliefs and to separate true social reasoning from generalized factual recall errors. Other work, including \textbf{ToMChallenges} \cite{ma2023tomchallenges}, has also emphasized that small changes in task wording, belief order, and information access can strongly affect measured ToM performance. This matters for OSCToM because observer-self conflict depends on the exact relation between what an agent knows, what the agent believes, and what the agent thinks another character believes. At the same time, large-scale behavioral evaluations were performed comparing LLM performance directly against human baselines \cite{becchio2024testing}, while multi-order assessments examined the limits of recursive depth \cite{street2024llms}. The results were mixed. LLMs sometimes matched adult human performance on fixed behavioral tasks, but they failed unpredictably when the task structure changed, suggesting weak general logical grounding \cite{shapira2023clever}. To address the limitations inherent in passive observation tasks, benchmarking efforts increasingly shifted toward scenarios involving information asymmetry and dynamic, interactive contexts. The introduction of the \textbf{FANToM} benchmark \cite{zheng2023fantom} was an important step in this regard, testing models in multi-party conversational settings where characters possess unequal access to ground-truth information. This change showed that even models excelling at static 1st-order scenarios drop sharply when forced to continuously update and track "who knows what" across an evolving dialogue \cite{chen2024tombench}. 

Alongside benchmark design, researchers have studied whether these successes and failures correspond to internal model representations. Recent interpretability work using techniques such as linear probing on hidden activations has provided early evidence that models form explicit internal representations of belief states for both self and others \cite{zhu2024language}. This creates an important distinction. A model may encode belief information internally, but it may still fail to use that information during complex inference or adversarial text-based reasoning. Seeking broader evaluation contexts beyond text, researchers have also introduced \textbf{OpenToM} \cite{xu2024opentom}, which emphasizes multi-modal and comprehensive video-based ToM tracking, further showing that models struggle to maintain mental-state coherence when information is distributed across complex or novel contexts. In addition, specialized benchmarks like \textbf{NegotiationToM} \cite{chan2024negotiation} have shown that when models are required to use ToM strategically, acting on inferred beliefs to achieve a goal rather than passively answering a question, their cognitive models often collapse. Taken together, these findings suggest that LLMs may contain some static reasoning components, but dynamic belief tracking in adversarial or socially complex settings remains difficult.

For this reason, recent ToM evaluation has moved toward adversarial and programmatic data generation. The \textbf{ExploreToM} framework \cite{sclar2024exploretom} is a leading example in this area, using a domain-specific language (DSL) guided by heuristic $A^*$ search to synthesize complex adversarial stories that programmatically manipulate informational access. By actively searching for scenarios that violate a model's existing heuristics, \textit{ExploreToM} successfully revealed large performance drops in models like Llama-3-70B. However, heuristic-based generation has computational scaling limits and a rigid search space, particularly when attempting to construct logically sound scenarios beyond 3rd-order recursive depth. More importantly, the existing adversarial literature focuses mainly on external perspective-taking, and leaves out the concept of \textbf{Observer-Self Conflict}, a state where an agent's recursive attribution of another's belief directly contradicts their own internal factual knowledge. Our work addresses this gap by replacing rigid heuristic search with a policy-driven Reinforcement Learning (RL) approach and a compositional surrogate evaluation pipeline. This allows us to model and generate Observer-Self Conflicts at scale, pushing ToM evaluation beyond linear tracing and toward human-like cognitive conflict.

A related line of work studies how training data can shape reasoning behavior after pretraining. Curriculum learning \cite{bengio2009curriculum} is relevant here because high-order ToM tasks are not uniform in difficulty. A model that cannot reliably solve first-order false-belief cases is unlikely to handle third- or fourth-order nested beliefs in a stable way. This is why recent work on reasoning benchmarks often separates simpler belief tracking from tasks that require recursive updates across several agents. For OSCToM, this observation motivates a staged training design: the model first sees lower-order belief conflicts and then moves to harder observer-self cases. This design follows the broader idea that reasoning ability can improve when examples are ordered by difficulty rather than presented as a single mixed dataset. If the story becomes inconsistent, then a wrong model answer may reflect confusion in the data rather than a real failure of ToM reasoning. This makes verification an important part of adversarial ToM generation.

These points clarify the position of OSCToM relative to prior work. Existing benchmarks show that LLMs struggle with recursive belief tracking, information asymmetry, and strategic use of mental-state information. Programmatic methods show that harder cases can be generated in a controlled way. OSCToM combines these directions by generating examples that are both adversarial and tied to a specific cognitive structure: the conflict between an observer's internal belief and the belief that the observer assigns to another agent. This focus is narrower than general social reasoning, but it allows the method to target a clear and difficult failure mode. 

\section{Methodology}
\begin{figure}[h]
    \centering
    \includegraphics[width=0.9\linewidth]{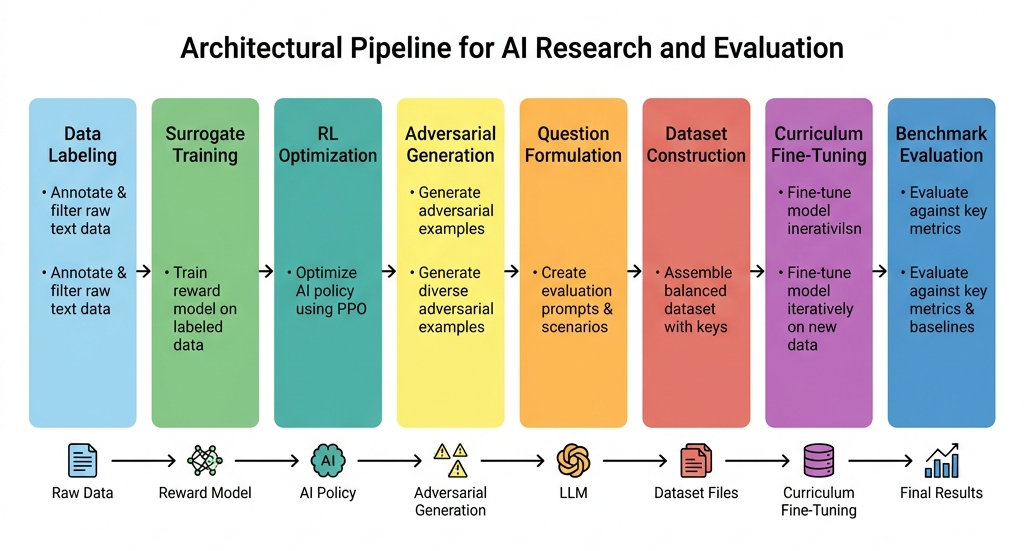}
    \caption{The 8-stage end-to-end OSCT verification and training workflow, mapping 
    the progression from initial surrogate distillation through to final model derivation.}
    \label{fig:workflow}
\end{figure}

OSCToM has four main components. First, we extend a domain-specific language so it can express Observer-Self Conflict. Second, we train lightweight surrogate evaluators from LLM-distilled difficulty annotations. These evaluators provide a low-cost estimate of narrative hardness. Third, a reinforcement learning agent uses the surrogate reward to search the DSL state space and select action sequences that create adversarial stories. Fourth, the generated dataset is used to fine-tune \textbf{Llama-3.1-8B-Instruct} through a two-stage curriculum that gradually introduces higher orders of recursive belief conflict. Figure~\ref{fig:workflow} shows the full pipeline.

\subsection{Extended Domain-Specific Language}
To move beyond the linear information flow of existing benchmarks like \textit{ExploreToM} \cite{sclar2024exploretom}, we introduce the \textbf{OSCT-DSL}. This language formalizes "Observer-Self Conflict," a state where an agent's recursive model of another's belief is in direct tension with their own factual knowledge. By expanding the state space to a 4th-order hierarchy, OSCT-DSL allows us to programmatically verify complex scenarios that were previously difficult to track.

We define a world state $\mathcal{W}$ and a belief state $B$ for any agent $i$ over a proposition $p$. While first-order beliefs $B_{i}^{(1)}(p)$ represent an agent's internal model of reality, we formalize higher-order recursive beliefs as:
\begin{equation}
    B_{i, j, \dots, n}^{(k)}(p) = B_{i}^{(1)}(B_{j, \dots, n}^{(k-1)}(p))
\end{equation}
where $k \in \{2, 3, 4\}$ denotes the order of recursion. This allows our framework to synthesize deep information asymmetries.

To generate these high-order conflicts, we move beyond simple physical movements by implementing a set of adversarial primitive actions designed to break informational symmetry. These include \textbf{Deceptive Localization} (\texttt{lie\_about\_location}), where an agent communicates a false proposition $p'$ to another, creating a first-order false belief $B_{j}^{(1)}(p')$ while the liar's internal model remains grounded in the true world state. Another operator is \textbf{Asymmetric Observation} (\texttt{one\_way\_mirror}), which allows an agent to witness an action without the observed party gaining a reciprocal second-order belief. This separates the shared-experience model common in simpler ToM benchmarks. Finally, the \textbf{Recursive Deception} (\texttt{double\_bluff}) is a composite action where one agent manipulates another into passing a lie to a third party, layering a third-order recursive belief on top of an initial first-order falsehood.

\subsection{Compositional Surrogate Model Training}

A major difficulty in applying Reinforcement Learning to natural language generation is the cost of computing rewards. During training, the RL agent must evaluate thousands of candidate story trajectories. Each trajectory requires a judgment about cognitive complexity. If every judgment is made by a large LLM, each query can require about 70 billion floating-point operations. This makes the optimization loop too expensive and prevents the agent from receiving a dense reward signal for policy improvement.

To reduce this cost, we implement the \textbf{Compositional Surrogate Evaluation Pipeline}. We use \textbf{Knowledge Distillation} \cite{hinton2015distilling} to transfer the evaluation behavior of \textbf{Llama-3.3-70B} \cite{dubey2024llama3} into six compact \textbf{DistilBERT} \cite{sanh2019distilbert} student modules trained on 10,000 annotated stories. DistilBERT is used as the student backbone because it retains 97\% of BERT's language understanding performance with 40\% fewer parameters. Its small size also allows all six modules to be loaded together as a GPU inference ensemble, avoiding model-swap overhead during RL training.

The reward is divided into \textit{six independent} cognitive dimensions instead of being reduced to one scalar score. This choice reflects the fact that ToM reasoning involves different processes \cite{frith2006neural}. For example, a story may contain many deceptive actions but still have shallow recursive depth. These two cases should guide the generator in different ways. Separate outputs therefore give the agent a more structured and interpretable reward signal. The six modules are defined as follows:

\begin{itemize}
    \item \textbf{False Belief Detector} — A binary classifier identifying 
    discrepancies between an agent's internal belief and the ground-truth world state. 
    Its presence is necessary: any story without a false belief is by definition 
    not a Theory of Mind scenario \cite{wimmer1983beliefs}.

    \item \textbf{ToM Depth Classifier} — A 4-class estimator classifying the 
    maximum order of recursive belief embedding present ($1^{\text{st}}$ through 
    $4^{\text{th}}$ order). It is given the second-highest reward weight, as recursive 
    depth is the most direct indicator of adversarial difficulty in OSCToM scenarios.

    \item \textbf{Deception Scorer} — A continuous scorer measuring the density of 
    manipulative operators (lies, double-bluffs, fake memory implants) normalized by 
    story length. Normalization prevents the agent from exploiting artificially 
    extended narratives as a reward hack.

    \item \textbf{Social Complexity Scorer} — Measures the frequency of inter-agent 
    communication events relative to the active cast of agents. Observer-Self Conflict 
    states cannot arise without rich multi-agent interaction; this module ensures the 
    generator is encouraged to construct dense social graphs.

    \item \textbf{Temporal Complexity Scorer} — Measures the number of 
    temporally distinct world-state transitions relevant to a target agent's final 
    belief state. This directly addresses the well-documented failure of language 
    models to track chronologically ordered belief changes \cite{zhu2023hitom}.

    \item \textbf{OSCT Detector} — A binary classifier with continuous confidence 
    scoring that specifically detects Observer-Self Conflict states: the condition 
    where an observer's recursive belief model directly contradicts their own 
    first-person knowledge of reality. As the primary optimization target of this 
    work, it receives the highest single reward weight in the composite function.
\end{itemize}

The composite reward signal is formulated as a weighted linear combination of five 
continuous surrogate outputs $S_k \in [0,1]$, while the False Belief Detector is used
as a hard validity constraint to reject non-ToM stories:
\begin{equation}
    R(\text{Story}) = 0.40 \cdot S_{\text{osct}} + 0.30 \cdot S_{\text{depth}} 
    + 0.15 \cdot S_{\text{dec}} + 0.075 \cdot S_{\text{soc}} 
    + 0.075 \cdot S_{\text{temp}}
    \label{eq:reward}
\end{equation}
The weights give priority to OSCT Detection and ToM Depth, which together account for 70\% of the reward. This keeps the policy focused on high-order Observer-Self Conflict states. The remaining 30\% is assigned to deception density, social complexity, and temporal complexity. These terms reduce the chance that the agent exploits only one reward dimension and help preserve narrative coherence. With this pipeline, per-story evaluation drops from about 14 seconds for an LLM query to less than 50 milliseconds with surrogate inference. This speedup makes it practical to construct a 15,000-sample adversarial corpus.

\subsection{Surrogate-Guided RL Training}
We model adversarial story generation as a Markov Decision Process (MDP), defined by the tuple $(\mathcal{S}, \mathcal{A}, P, R, \gamma)$. The state and action spaces come from the Extended DSL. The main challenge is to find sequences of cognitive and social operators, such as \texttt{double\_bluff} or \textit{asymmetric observation}, that increase both narrative conflict and recursive depth. To choose the generation policy, we compared several reinforcement learning methods: Asynchronous Advantage Actor-Critic (A2C) \cite{mnih2016asynchronous}, Proximal Policy Optimization (PPO) \cite{schulman2017ppo}, and Deep Q-Networks (DQN) \cite{mnih2015human}.

Table \ref{tab:rl_comparison} summarizes the comparison. We selected Deep Q-Networks (DQN) as the main generation policy because it learned the discrete symbolic transitions of the OSCT-DSL more efficiently. Policy-gradient methods such as A2C and Recurrent PPO were less stable during reward convergence. In contrast, the DQN generator maintained more stable value estimates for DSL states. This made it better suited for finding ToM failure cases that heuristic search methods may overlook \cite{sclar2024exploretom}.

\begin{figure}[h]
    \centering
    \includegraphics[width=0.65\textwidth]{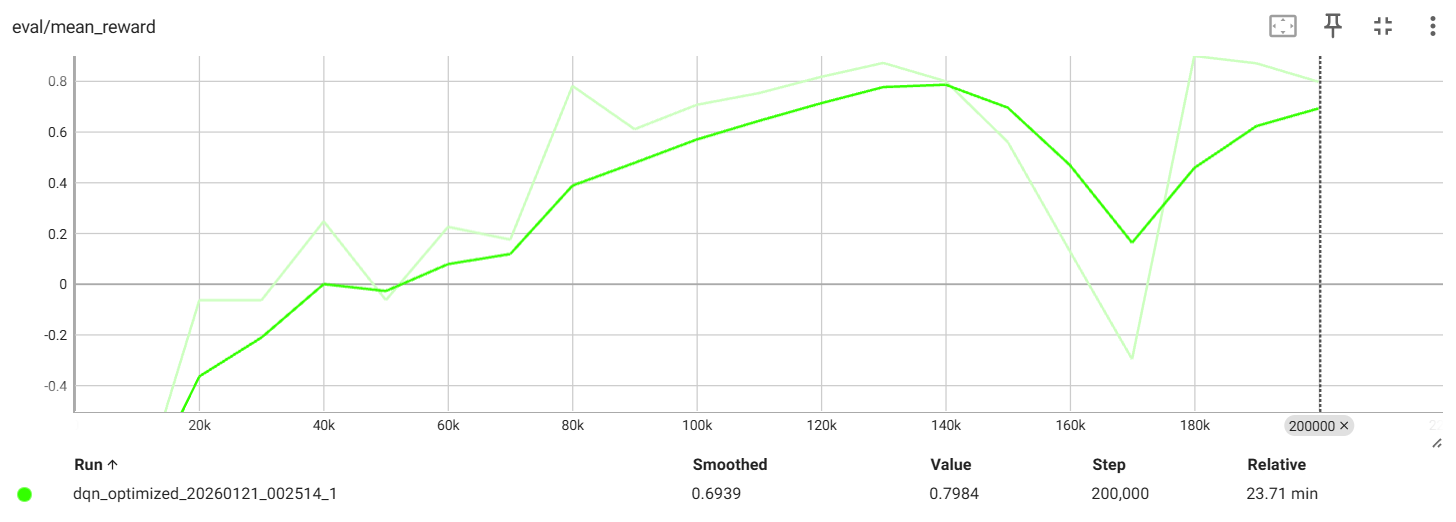}
    \caption{Learning curve for the optimized DQN generator. The vertical axis represents the mean episodic reward derived from the surrogate ensemble across 200,000 timesteps of adversarial training.}
    \label{fig:dqn_learning_curve}
\end{figure}

Figure \ref{fig:dqn_learning_curve} shows the training progress of our optimized DQN architecture in Figure \ref{fig:dqn_learning_curve}. Using a replay buffer of $100,000$ transitions and a target update interval of $500$ steps, the model converged stably and consistently produced complex 4th-order recursive scenarios. As shown in the performance summary, the DQN-based generator achieved the best balance between mean narrative hardness ($0.515$) and training efficiency ($16$m $37$s), making it the most suitable engine for our large-scale data synthesis pipeline.

\begin{table}[h]
\centering
    \caption{Comparative performance of reinforcement learning architectures. DQN was chosen as the primary generator for the OSCToM framework due to its superior stability and output quality.}
\label{tab:rl_comparison}
\begin{tabular}{@{}lcccc@{}}
\toprule
\textbf{Algorithm} & \textbf{Mean Reward} & \textbf{Mean Hardness} & \textbf{Stability} & \textbf{Training Time} \\ \midrule
A2C           & -0.060               & 0.519                  & Low                  & 25m 08s                \\
PPO           & 0.900                & 0.517                  & Medium               & 22m 53s                \\
Recurrent PPO & -0.060               & 0.518                  & Low                  & 56m 36s                \\
\textbf{DQN}  & \textbf{0.900}       & \textbf{0.515}         & \textbf{High}        & \textbf{16m 37s}       \\ \bottomrule
\end{tabular}
\end{table}

\subsection{Adversarial Dataset Generation}
The OSCT corpus is produced through a four-stage pipeline that combines symbolic trajectory optimization with LLM-based narrative enhancement. At the start of each episode, \texttt{ToMStoryEnv} samples a context from predefined agent names, room layouts, and object inventories. The trained DQN policy then rolls out over 15 discrete DSL operations. It selects physical and cognitive actions, including \texttt{lie\_about\_location}, \texttt{double\_bluff}, \texttt{one\_way\_mirror\_observation}, and \texttt{fake\_memory\_implant}, to maximize the composite hardness reward. At each timestep, the policy receives a 256-dimensional observation vector encoding the normalized story length, active agent and object counts, and the surrogate module scores from the preceding episode.

The terminal reward is computed by the compositional surrogate ensemble across five continuous cognitive dimensions. The False Belief Detector acts as a validity gate. The weights emphasize Observer-Self Conflict detection and recursive depth:
\begin{equation}
    H = 0.40 \cdot S_{\text{osct}} + 0.30 \cdot S_{\text{depth}} +
        0.15 \cdot S_{\text{dec}} + 0.075 \cdot S_{\text{soc}} +
        0.075 \cdot S_{\text{temp}}, \quad H \in [0, 1]
\end{equation}
A three-phase curriculum scheduler changes the relative weights of hardness, diversity, and validity during training. Early episodes emphasize structural validity. Later episodes place more weight on adversarial hardness. The final symbolic script is passed to \textbf{Llama-3.3-70B} \cite{dubey2024llama3} through the OpenRouter API. This step converts the DSL trace into natural prose while preserving the intended cognitive markers, including deception chains, information asymmetry, and non-linear temporal ordering. The target \texttt{overall\_hardness} score is set above 0.85.

\begin{figure}[h]
    \centering
    \begin{minipage}{0.48\textwidth}
        \centering
        \includegraphics[width=\linewidth]{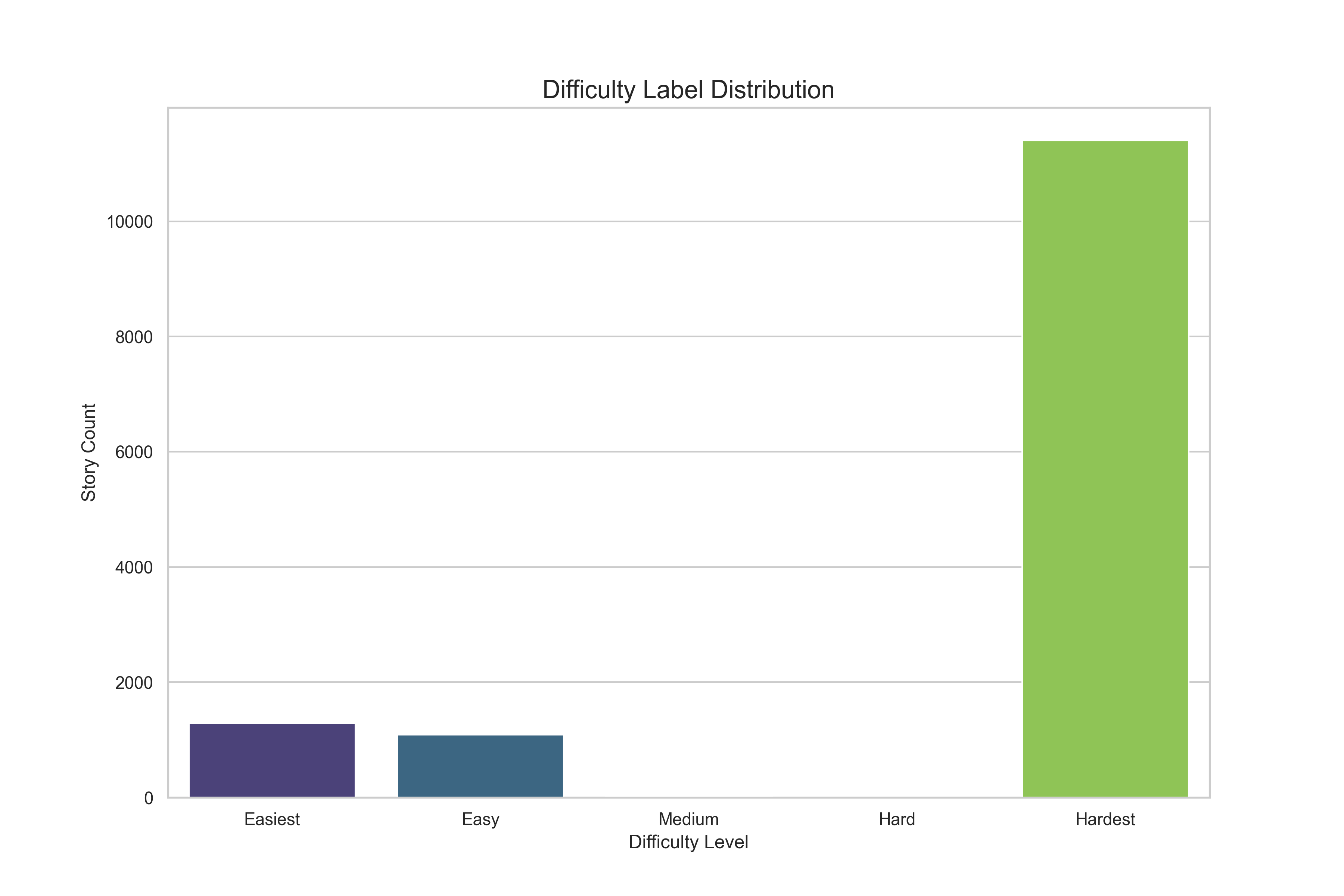}
        \caption{Difficulty tier distribution across the OSCT corpus.}
        \label{fig:difficulty_dist}
    \end{minipage}\hfill
    \begin{minipage}{0.48\textwidth}
        \centering
        \includegraphics[width=\linewidth]{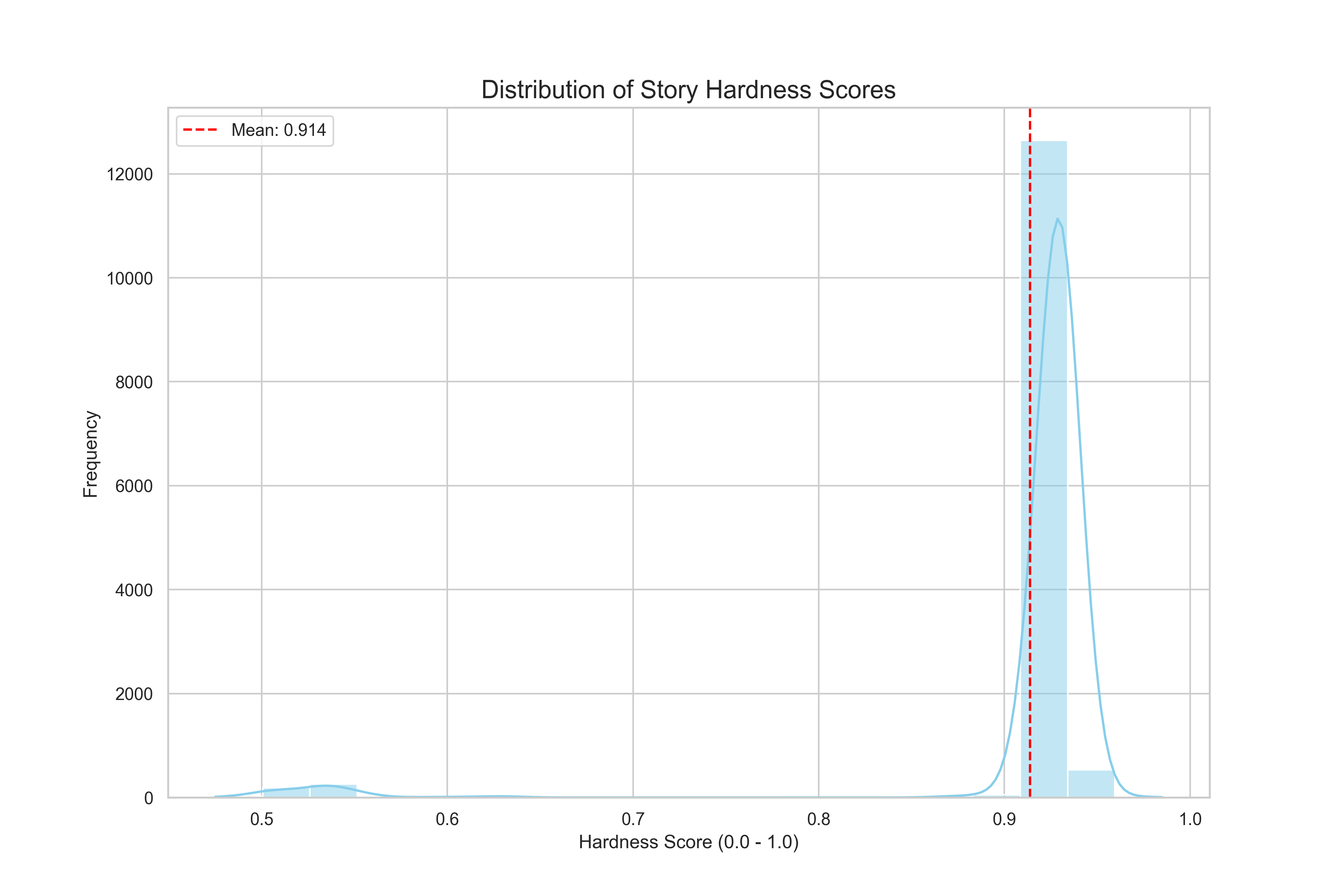}
        \caption{Histogram of aggregate hardness scores across the 15,000-sample corpus.}
        \label{fig:hardness_hist}
    \end{minipage}
\end{figure}

\begin{figure}[h]
    \centering
    \includegraphics[width=0.6\textwidth]{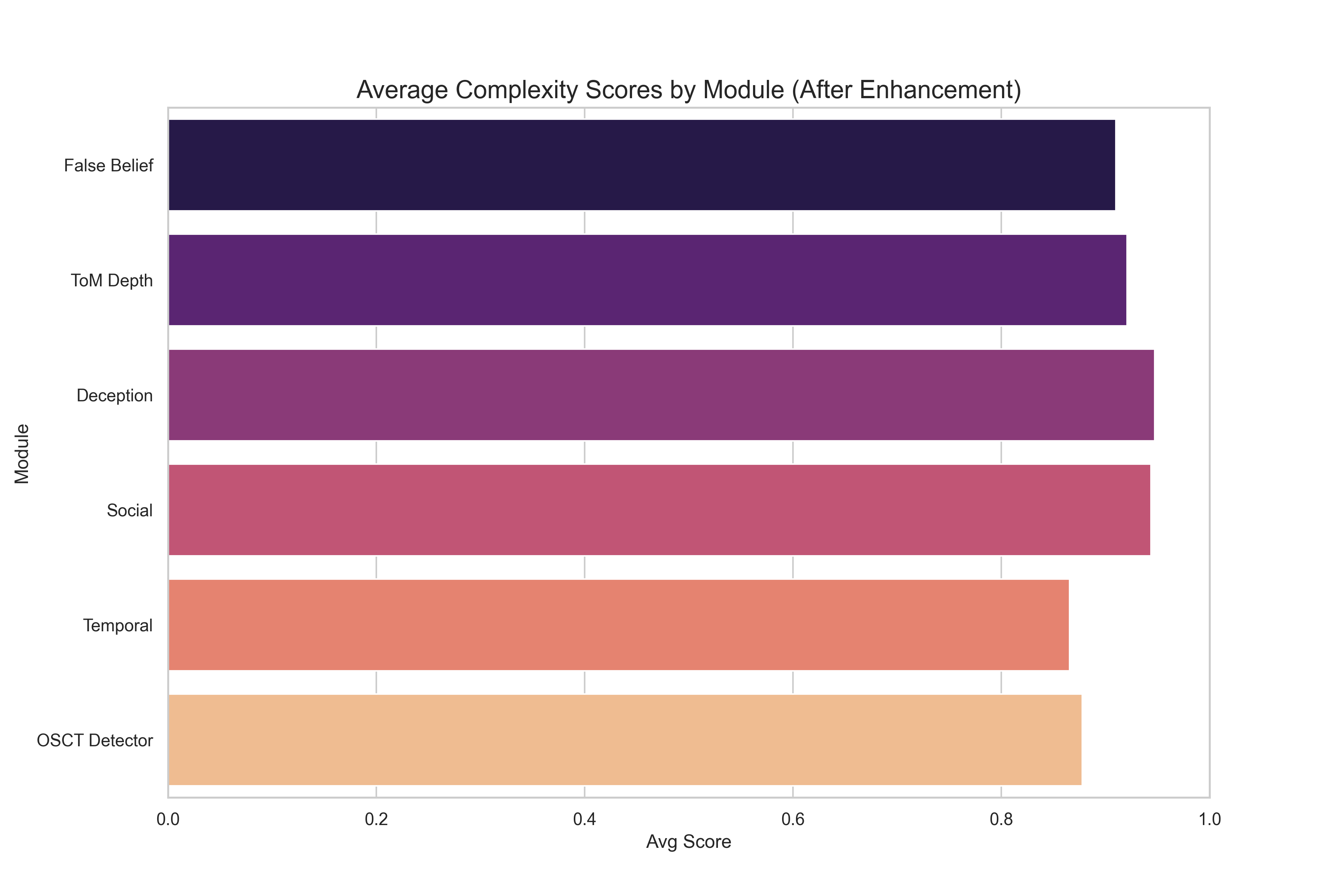}
    \caption{Per-dimension surrogate complexity scores across the six cognitive modules.}
    \label{fig:module_complexity}
\end{figure}

The \texttt{ToMQuestionGenerator} then reads the internal belief dictionaries in the \texttt{ExtendedToMDSL} state. It extracts up to five question-answer pairs per story for each recursion order, from first-order to fourth-order belief queries. After generation, each story receives a difficulty label from one of five tiers. The tiers are assigned using percentile thresholds ($P_{20}, P_{40}, P_{60}, P_{80}$) computed from the final hardness-score distribution. Figures~\ref{fig:difficulty_dist}, \ref{fig:hardness_hist}, and~\ref{fig:module_complexity} show the resulting difficulty distribution and surrogate scores for the 15,000-sample corpus.

\subsection{Two-Stage Curriculum Fine-Tuning}
The final component of OSCToM is supervised fine-tuning of \textbf{Llama-3.1-8B-Instruct} \cite{dubey2024llama3} on the OSCT corpus. We do not train on the full dataset in one pass. Instead, we use a two-stage curriculum strategy \cite{bengio2009curriculum}, implemented in the \texttt{CurriculumTrainer} class. The motivation is simple: high-order reasoning is much harder than low-order false-belief reasoning. If the model sees 4th-order Observer-Self Conflict examples too early, training becomes less stable and generalization decreases.

In \textbf{Stage 1 (Foundation)}, the model is trained only on stories
containing questions at the 1st and 2nd order of Theory of Mind recursion, automatically
filtered from the full OSCT dataset by selecting entries where at least one associated
question carries a \texttt{tom\_order} value of 2 or below. This stage grounds the
model in basic perspective-taking and false-belief resolution before deceptive or
asymmetric constructs are introduced. In \textbf{Stage 2 (Mastery)}, the full OSCT
corpus is used, including the complete range of 3rd and 4th-order recursive
scenarios. The checkpoint produced by Stage 1 is used as the initialization point for
Stage 2, ensuring that foundational belief-tracking capabilities are preserved rather
than overwritten during the higher-order optimization.

\begin{figure}[h]
    \centering
    \includegraphics[width=0.7\linewidth]{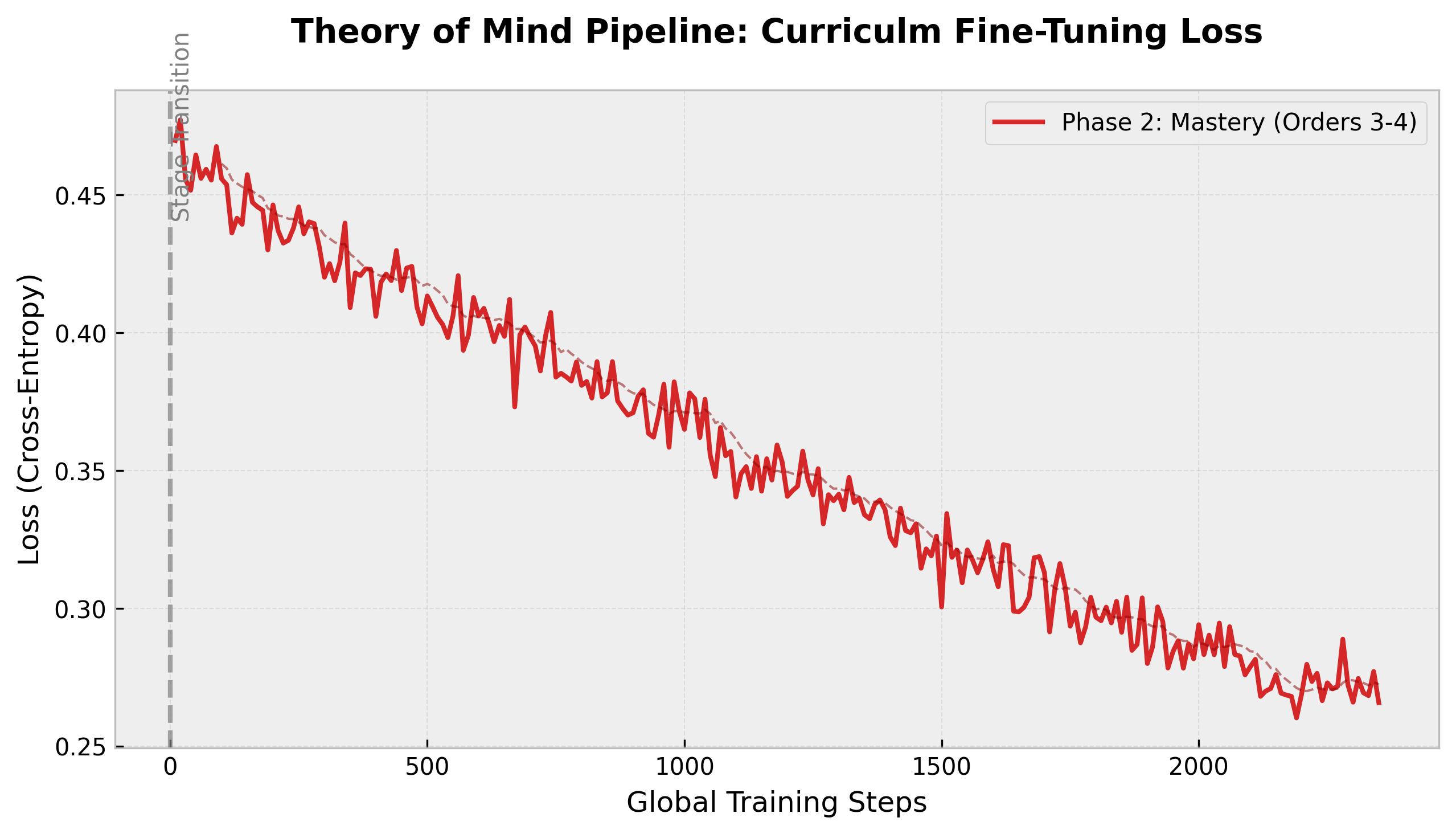}
    \caption{Training loss across both curriculum stages of the OSCToM-8B fine-tuning pipeline. Stage 1 optimizes on 1st and 2nd-order ToM scenarios, while Stage 2 extends training to the full adversarial OSCT corpus containing 3rd and 4th-order
    recursive conflicts.}
    \label{fig:training_loss}
\end{figure}

Parameter-efficient fine-tuning is performed through \textbf{Low-Rank Adaptation
(LoRA)} \cite{hu2022lora}, applied across all seven projection layers of the attention
and feed-forward blocks (\texttt{q\_proj}, \texttt{k\_proj}, \texttt{v\_proj},
\texttt{o\_proj}, \texttt{gate\_proj}, \texttt{up\_proj}, \texttt{down\_proj}).
The adapter is configured with rank $r = 16$, scaling factor $\alpha = 32$, and dropout
$= 0.05$. The base model is loaded in 4-bit quantization via the Unsloth optimization
framework \cite{han2023unsloth}, allowing training within a single-GPU budget.
Each stage uses a per-device batch size of 2 with 8 gradient accumulation steps
(effective batch size of 16), a linear learning rate scheduler, and the AdamW-8bit
optimizer. Training instances are formatted as structured prompts of the form
\textit{Story} $\rightarrow$ \textit{Question} $\rightarrow$ \textit{Answer}, with a
maximum sequence length of 1,024 tokens. The training loss across both
curriculum stages is shown in Figure~\ref{fig:training_loss}.

\section{Results and Evaluation}

We evaluate OSCToM-8B on benchmark accuracy and inference efficiency. This distinction is important because ExploreToM \cite{sclar2024exploretom} reports results from iterative A* heuristic search rather than from direct model inference. That procedure has an average latency of 15.0 seconds per query and inference complexity $\mathcal{O}(N)$. OSCToM-8B uses single-pass neural inference. Its measured latency is 2.62 seconds, giving a \textbf{5.7x} reduction in response time while exceeding ExploreToM's FANToM accuracy by 378 percentage points.

\subsection{Benchmark Accuracy}

We evaluate OSCToM-8B against seven baseline systems across four Theory of Mind
benchmarks: \textbf{ToMi} \cite{le2019tomi}, \textbf{Hi-ToM} \cite{zhu2023hitom},
\textbf{BigToM} \cite{gandhi2023bigtom}, and \textbf{FANToM} \cite{zheng2023fantom}.
The baselines include Llama-3.1-8B-Base \cite{dubey2024llama3}, Mistral-NeMo-12B
\cite{mistralnemo2024}, Phi-3-Medium-14B \cite{abdin2024phi3}, Qwen2.5-14B and
Qwen2.5-32B \cite{qwen252024technicalreport}, Gemma-2-27B \cite{gemma_2_2024}, and
the reported results of ExploreToM \cite{sclar2024exploretom}. Full results are
shown in Table~\ref{tab:benchmark}.

\begin{table}[h]
\centering
\caption{Accuracy (\%) comparison across four Theory of Mind benchmarks.
\textbf{Bold} denotes the best result per column.
$^\dagger$ denotes results as reported in the original paper, produced via
iterative A* heuristic search rather than direct model inference.}
\label{tab:benchmark}
\setlength{\tabcolsep}{8pt}
\renewcommand{\arraystretch}{1.35}
\begin{tabular}{@{}lrcccc@{}}
\toprule
\textbf{Model} & \textbf{Params} & \textbf{ToMi} & \textbf{Hi-ToM} & \textbf{BigToM} & \textbf{FANToM} \\
\midrule
ExploreToM$^\dagger$ \cite{sclar2024exploretom} & 8B  & 95.0 & 59.0 & 81.0 & 0.2 \\
Llama-3.1-8B-Base \cite{dubey2024llama3}         & 8B  & 64.3 & 67.8 & 86.0 & 66.0 \\
Mistral-NeMo-12B \cite{mistralnemo2024}           & 12B & 66.8 & 25.6 & \textbf{90.5} & 37.5 \\
Phi-3-Medium-14B \cite{abdin2024phi3}             & 14B & 76.0 & 65.3 & 85.5 & 51.0 \\
Qwen2.5-14B \cite{qwen252024technicalreport}      & 14B & 53.8 & 57.9 & 29.5 & 54.5 \\
Qwen2.5-32B \cite{qwen252024technicalreport}      & 32B & 74.7 & \textbf{68.5} & 47.5 & 46.5 \\
Gemma-2-27B \cite{gemma_2_2024}                   & 27B & 53.3 & 20.6 & 18.5 & 38.5 \\
\midrule
\textbf{OSCToM-8B (Ours)}                         & \textbf{8B} & \textbf{79.5} & 65.3 & 89.8 & \textbf{76.0} \\
\bottomrule
\end{tabular}
\end{table}

\begin{figure}[h]
    \centering
    \includegraphics[height=8.5cm, keepaspectratio]{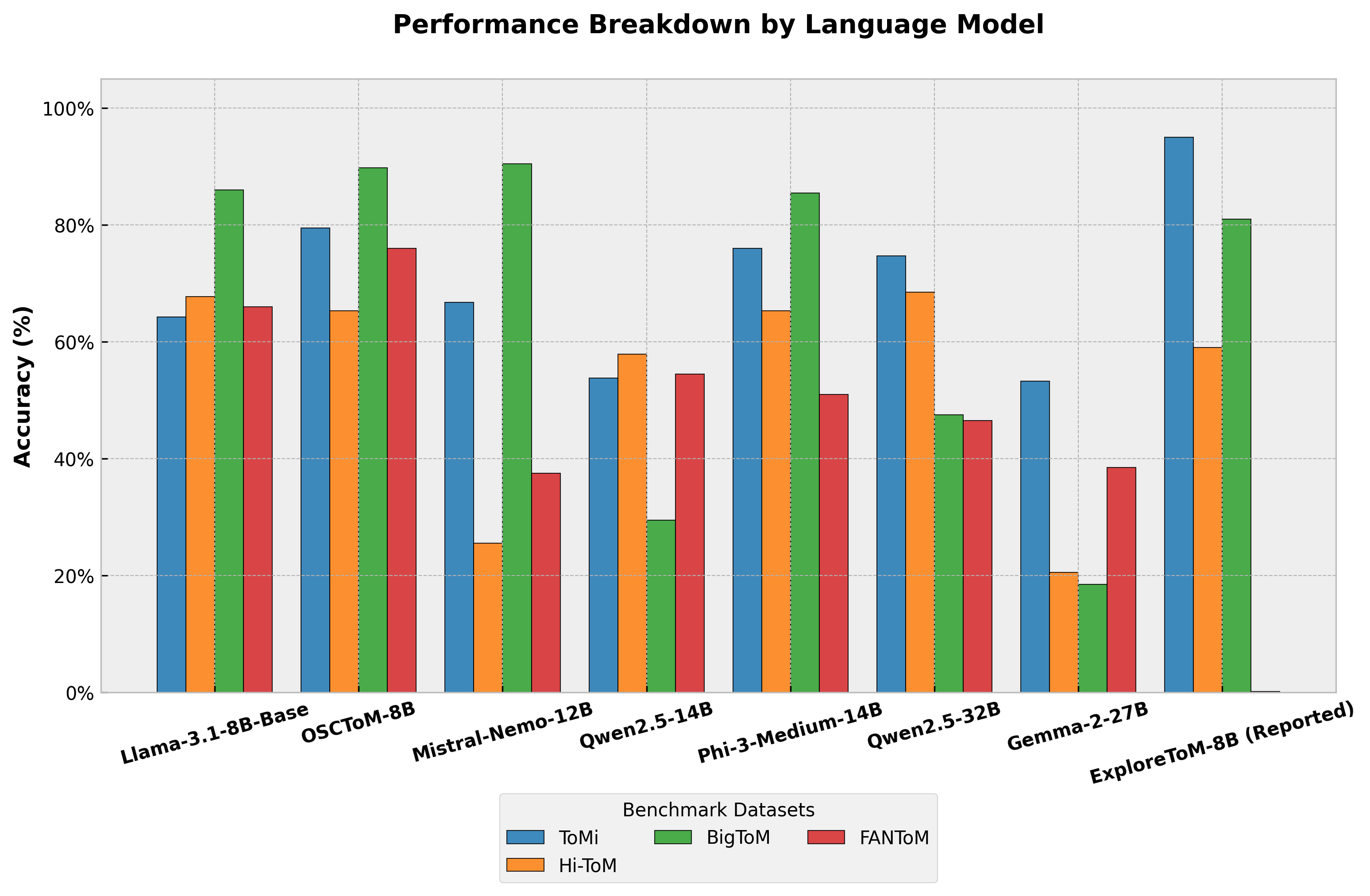}
    \caption{Bar chart comparison of benchmark accuracy across all evaluated models on ToMi, Hi-ToM, BigToM, and FANToM. OSCToM-8B achieves the best overall accuracy profile despite its smaller parameter count.}
    \label{fig:model_comparison}
\end{figure}

OSCToM-8B gives the highest accuracy on \textbf{ToMi} (79.5\%) and \textbf{FANToM} (76.0\%). It also exceeds several models with three to four times more parameters. FANToM is especially important because it uses multi-party conversations in which characters have asymmetric access to the true state of the world \cite{zheng2023fantom}. A model must update nested belief states as the dialogue changes. ExploreToM reports 0.2\% accuracy on FANToM, while OSCToM-8B reaches 76.0\%. This difference suggests a limitation of A*-guided heuristic synthesis: optimizing for the amount of information does not necessarily produce training data that teaches the model to reason about divergent nested beliefs in dynamic conversations. Figure~\ref{fig:model_comparison} gives the visual comparison across all models.

\begin{figure}[h]
    \centering
    \includegraphics[height=7.5cm, keepaspectratio]{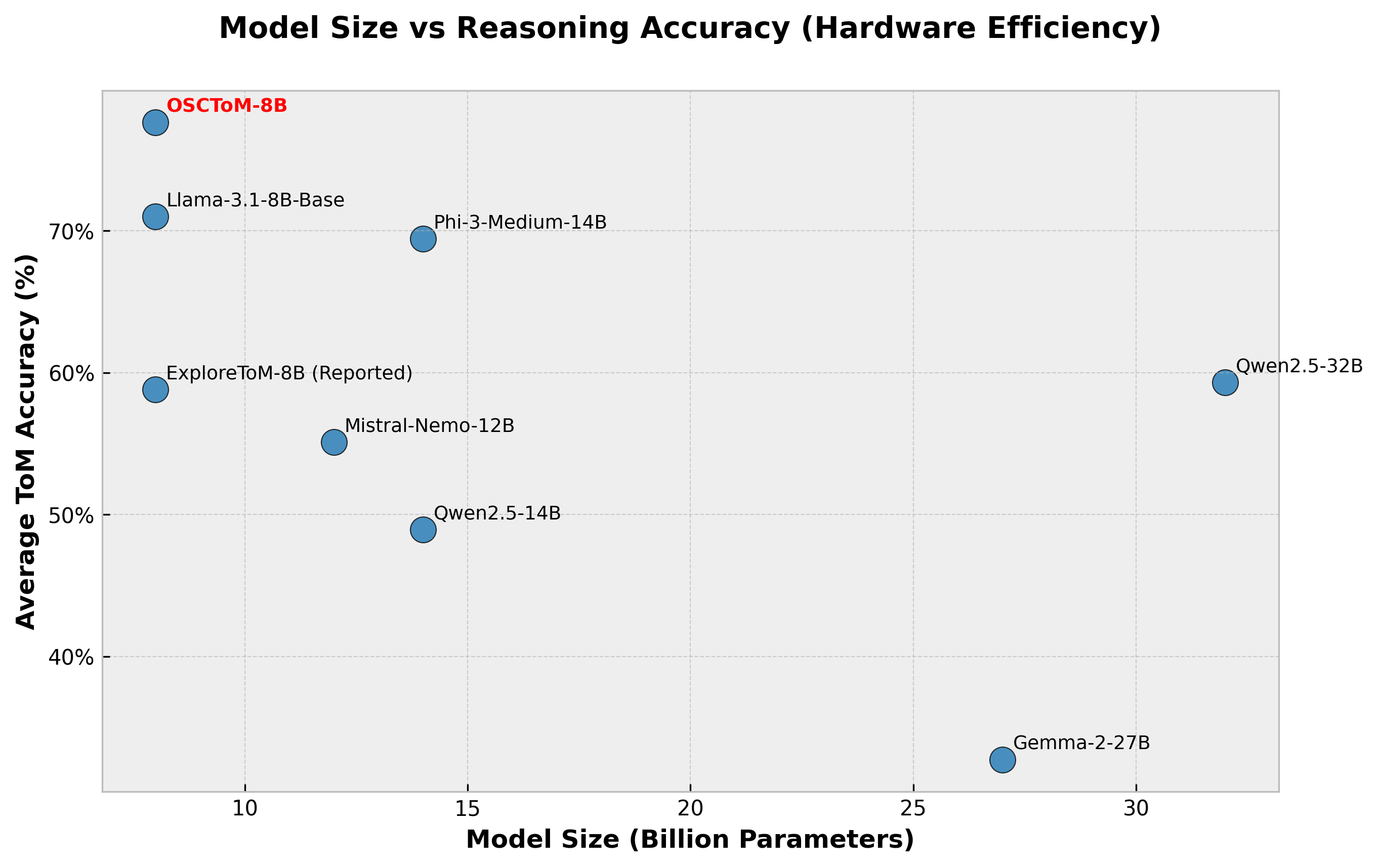}
    \caption{Scatter plot of model size (parameters) versus mean benchmark accuracy. OSCToM-8B shows the best accuracy-to-size ratio among all evaluated models.}
    \label{fig:efficiency_scatter}
\end{figure}

On \textbf{BigToM}, OSCToM-8B reaches 89.8\%, close to Mistral-NeMo-12B at 90.5\%, despite having 4 billion fewer parameters. On \textbf{Hi-ToM}, OSCToM-8B reaches 65.3\%, matching Phi-3-Medium-14B and exceeding Gemma-2-27B (20.6\%) and Mistral-NeMo-12B (25.6\%). These results support the value of the two-stage curriculum, which exposes the model to recursive belief conflict in increasing order of difficulty.

\subsection{Inference Efficiency}
A second important dimension of the evaluation concerns inference efficiency.
Current ToM systems such as ExploreToM rely on iterative A* search, which requires multiple sequential LLM calls per query to construct and validate an answer. This procedure has an inference complexity of $\mathcal{O}(N)$, where $N$ is the number of search steps, and results in an average response latency of 15.0 seconds per query. OSCToM-8B operates through \textit{direct neural inference}---a single forward pass through the fine-tuned model---achieving $\mathcal{O}(1)$ inference complexity and a measured average latency of 2.62 seconds. This architectural difference is summarized in Table~\ref{tab:inference_comparison}.

\begin{table}[h]
\centering
\caption{Inference characteristics of ExploreToM versus OSCToM-8B. The A* search
method used by ExploreToM is iterative and hardware-intensive; OSCToM-8B
performs constant-time single-pass inference.}
\label{tab:inference_comparison}
\setlength{\tabcolsep}{12pt}
\renewcommand{\arraystretch}{1.4}
\begin{tabular}{@{}lcc@{}}
\toprule
\textbf{Feature} & \textbf{ExploreToM} \cite{sclar2024exploretom} & \textbf{OSCToM-8B (Ours)} \\
\midrule
Inference Method  & A* Search + LLM             & Direct Neural Inference \\
Complexity        & $\mathcal{O}(N)$ (Iterative) & $\mathcal{O}(1)$ (Constant Time) \\
Avg. Latency      & 15.0s (Reported)            & 2.62s (Measured) \\
Hardware Load     & High (Multiple Passes)      & Low (Single Pass) \\
Real-time Use     & Not Recommended             & Fully Optimized \\
\bottomrule
\end{tabular}
\end{table}

\begin{figure}[h]
    \centering
    \includegraphics[width=0.45\linewidth]{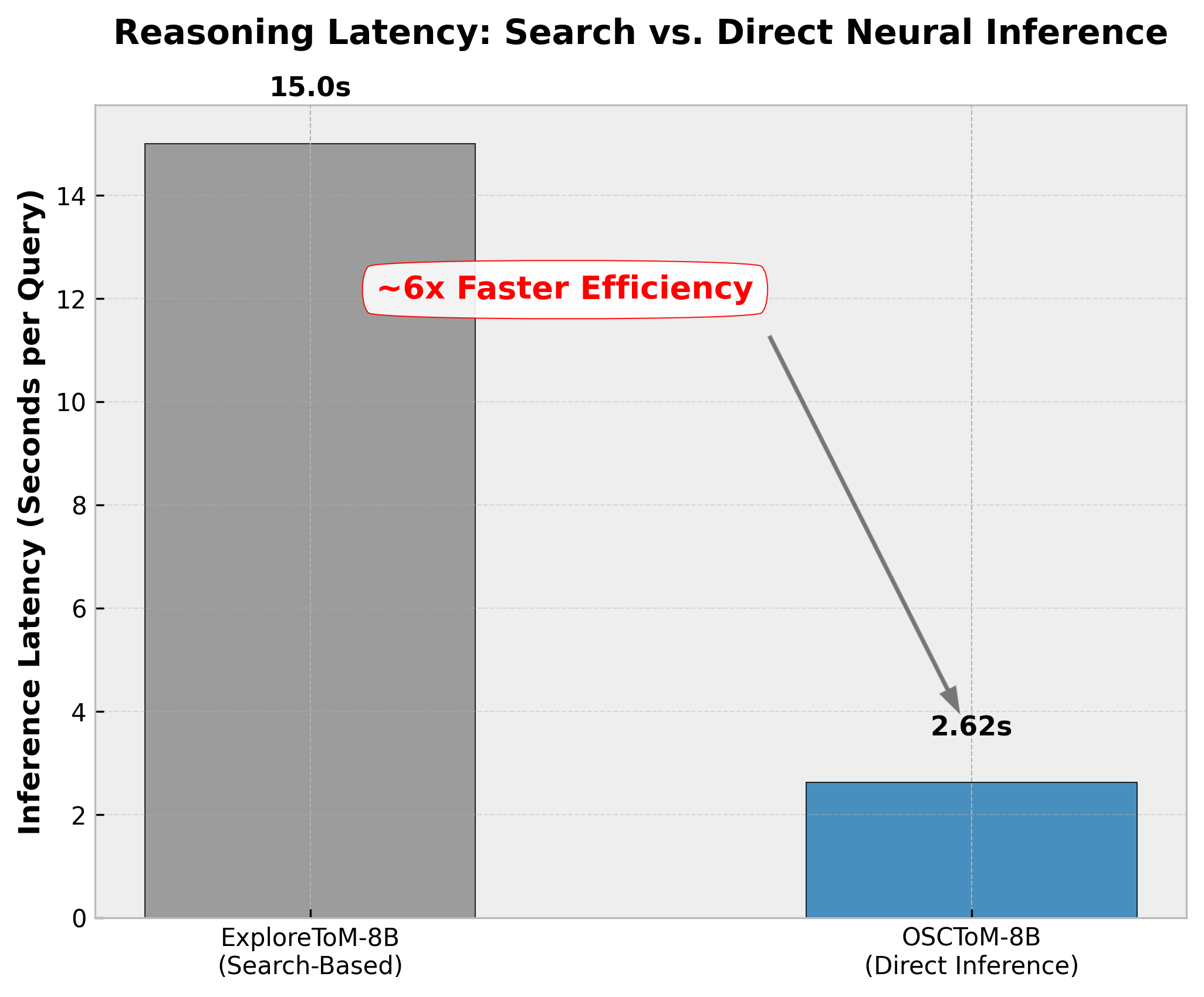}
    \caption{Inference throughput comparison across models. OSCToM-8B achieves
    a 5.7x reduction in average latency relative to ExploreToM's A* search
    procedure while maintaining superior accuracy.}
    \label{fig:inference_efficiency}
\end{figure}

Figures~\ref{fig:inference_efficiency} and~\ref{fig:efficiency_scatter} summarize the efficiency results. Figure~\ref{fig:inference_efficiency} compares throughput with benchmark accuracy and shows that OSCToM-8B avoids the iterative overhead of search-based methods. Figure~\ref{fig:efficiency_scatter} compares model size with aggregate benchmark accuracy. In this view, OSCToM-8B falls in the best region, with the highest combined accuracy per parameter count among the evaluated systems. These results suggest that adversarial training data and curriculum fine-tuning can improve both capability and computational practicality.

\subsection{Discussion and Limitations}

The results indicate that OSCToM is most useful when a task requires a model to keep track of several belief states at the same time. This is different from simple false-belief tests, where the main difficulty is often to identify what one character does or does not know. In Observer-Self Conflict scenarios, the model must also separate an observer's own belief from the belief that the observer attributes to another agent. This separation is the main reason the generated stories are useful for high-order ToM training. It gives the model repeated exposure to cases where the surface narrative is not enough and where the answer depends on the structure of nested mental states.

There are still limitations. The current DSL focuses mainly on belief conflicts built from location, observation, communication, and deception events. These operators are useful for controlled ToM evaluation, but they do not cover all forms of social reasoning. Human social understanding also includes emotion, uncertainty, memory, intention, trust, and moral judgment. Another limitation is that the generated stories are text-only. In real settings, beliefs may depend on visual cues, gestures, timing, or shared physical context. Extending OSCToM to those settings would require new state representations and new verification rules.

Finally, the surrogate pipeline makes generation efficient, but it also introduces a dependency on the quality of the surrogate labels. If the surrogate modules assign high scores to stories that are complex in form but weak in actual reasoning content, the generator may learn patterns that look difficult without being genuinely useful. For this reason, future versions of the framework should include stronger human or model-based audits of generated stories, especially for the highest difficulty tiers. These limitations do not change the reported results, but they clarify where the framework is strongest and where further validation is needed.

\section{Conclusion}

This paper introduced \textbf{OSCToM}, a framework for generating and training on adversarial Theory of Mind scenarios based on Observer-Self Conflict. This conflict occurs when an agent's recursive model of another agent's belief contradicts the agent's own internal knowledge. OSCToM combines an Extended DSL, a compositional surrogate pipeline, a DQN-guided RL generator, and a two-stage curriculum fine-tuning strategy for Llama-3.1-8B-Instruct \cite{dubey2024llama3}. The resulting OSCToM-8B model reaches \textbf{76\%} accuracy on FANToM \cite{zheng2023fantom}, compared with the \textbf{0.2\%} reported by ExploreToM \cite{sclar2024exploretom}. It also matches or exceeds models with up to four times more parameters. In addition, OSCToM-8B reduces inference latency by \textbf{5.7x} relative to A*-based pipeline approaches, making it a practical system for high-order social reasoning.

Future work will extend the DSL beyond locative belief conflicts to include emotional and epistemic states. We also plan to develop a dedicated OSCT evaluation benchmark and study multi-modal settings where belief states are distributed across visual and conversational evidence \cite{xu2024opentom}. Overall, these results suggest that principled adversarial data construction can be an efficient path toward more generalizable social intelligence in large language models.

\newpage
\bibliographystyle{unsrt}  
\bibliography{references}  %%% Remove comment to use the external .bib file (using bibtex).
%%% and comment out the ``thebibliography'' section.
\newpage

\appendix
\section{Appendix}
\subsection{Hyperparameter Tuning of the DQN Generator}
\label{app:hyperparameter}

Hyperparameter optimization was performed using the \textbf{Optuna} framework \cite{akiba2019optuna} with a Tree-structured Parzen Estimator (TPE) sampler across
25 trials, each evaluated over 50,000 timesteps and 10 test episodes. The search
spanned 11 parameters including learning rate, buffer size, discount factor $\gamma$,
soft update coefficient $\tau$, and the $\varepsilon$-greedy exploration schedule.
Of the 25 trials, 9 returned $-\infty$ reward due to policy divergence. Analysis of
these failures shows a consistent pattern: all divergent trials combined a small
replay buffer ($\leq 10{,}000$ transitions) with a high learning rate, causing the
Q-function estimates to become unstable before sufficient experience had accumulated.

\begin{figure}[h]
    \centering
    \includegraphics[width=0.75\linewidth]{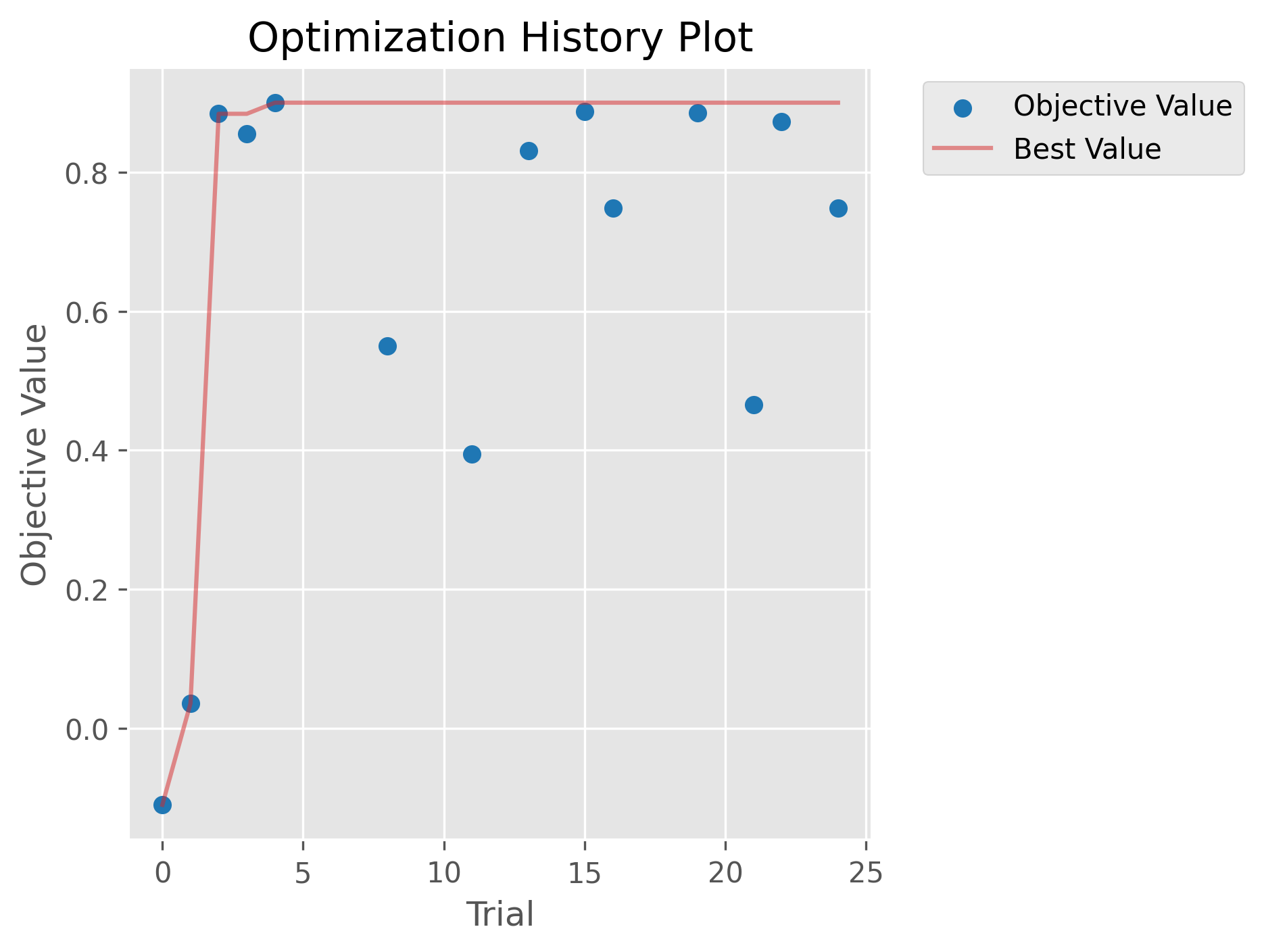}
    \caption{Optuna optimization history. The optimal reward of 0.900 was reached
    at Trial 4 and confirmed by subsequent trials, indicating successful convergence
    of the TPE sampler.}
    \label{fig:opt_history}
\end{figure}

The best configuration (Trial 4) specifies a learning rate of $5.95 \times 10^{-4}$,
buffer size of $100{,}000$, $\tau = 0.019$, $\gamma = 0.902$, and train frequency of
8 steps with 5 gradient updates per step. The especially low $\tau$ value ensures the
target network updates gradually, preventing Q-value overestimation in the
non-stationary surrogate-reward environment. Fanova importance analysis shows that
buffer size is the single most deterministic factor in training stability, with all
high-reward trials ($> 0.85$) sharing a buffer of $100{,}000$ regardless of other
parameter values.

\begin{figure}[h]
    \centering
    \begin{minipage}{0.48\textwidth}
        \centering
        \includegraphics[width=\linewidth]{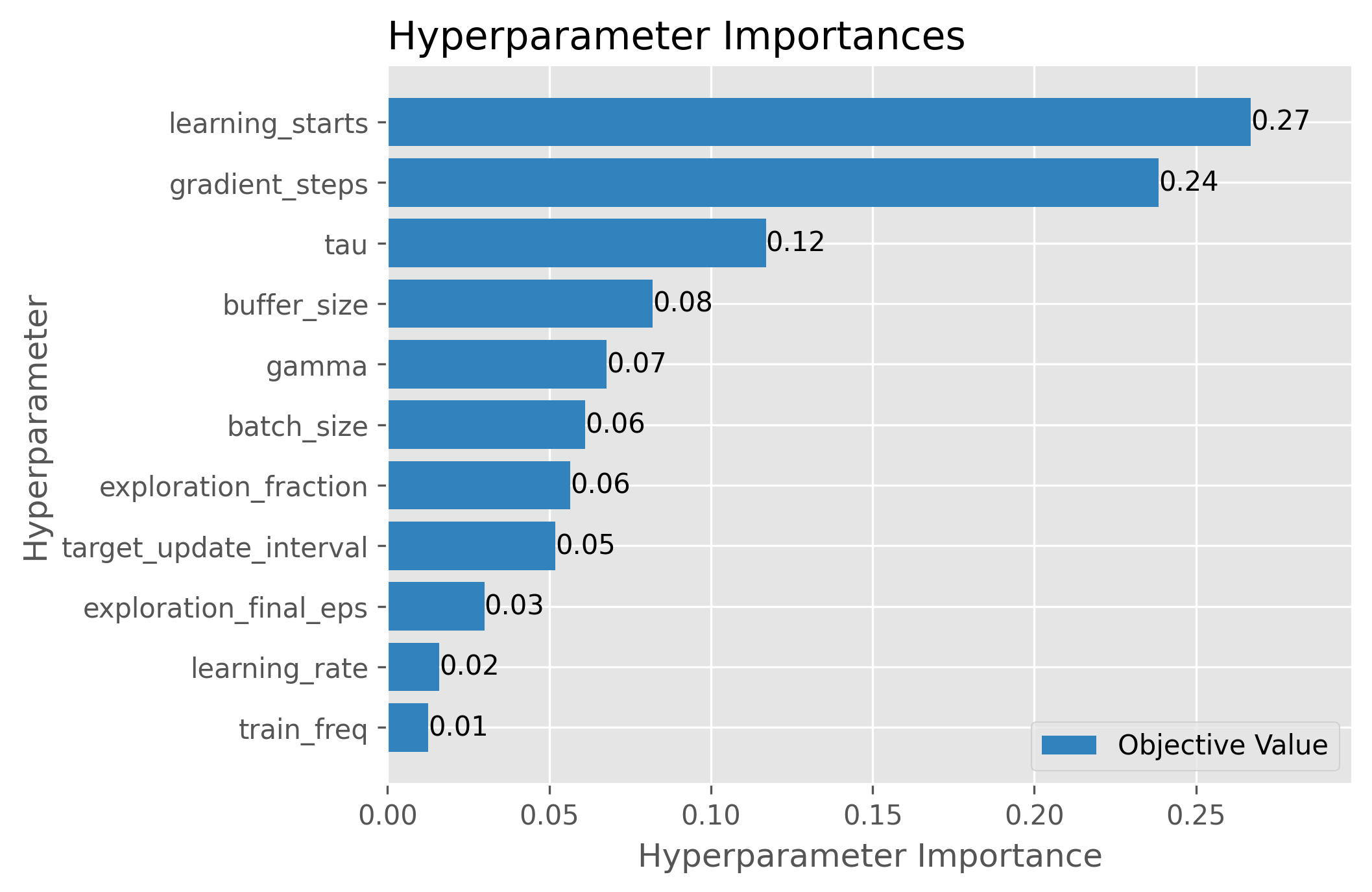}
        \caption{Fanova-estimated hyperparameter importance \cite{hutter2014efficient}. Buffer size and learning rate are dominant.}
        \label{fig:param_importance}
    \end{minipage}\hfill
    \begin{minipage}{0.48\textwidth}
        \centering
        \includegraphics[width=\linewidth]{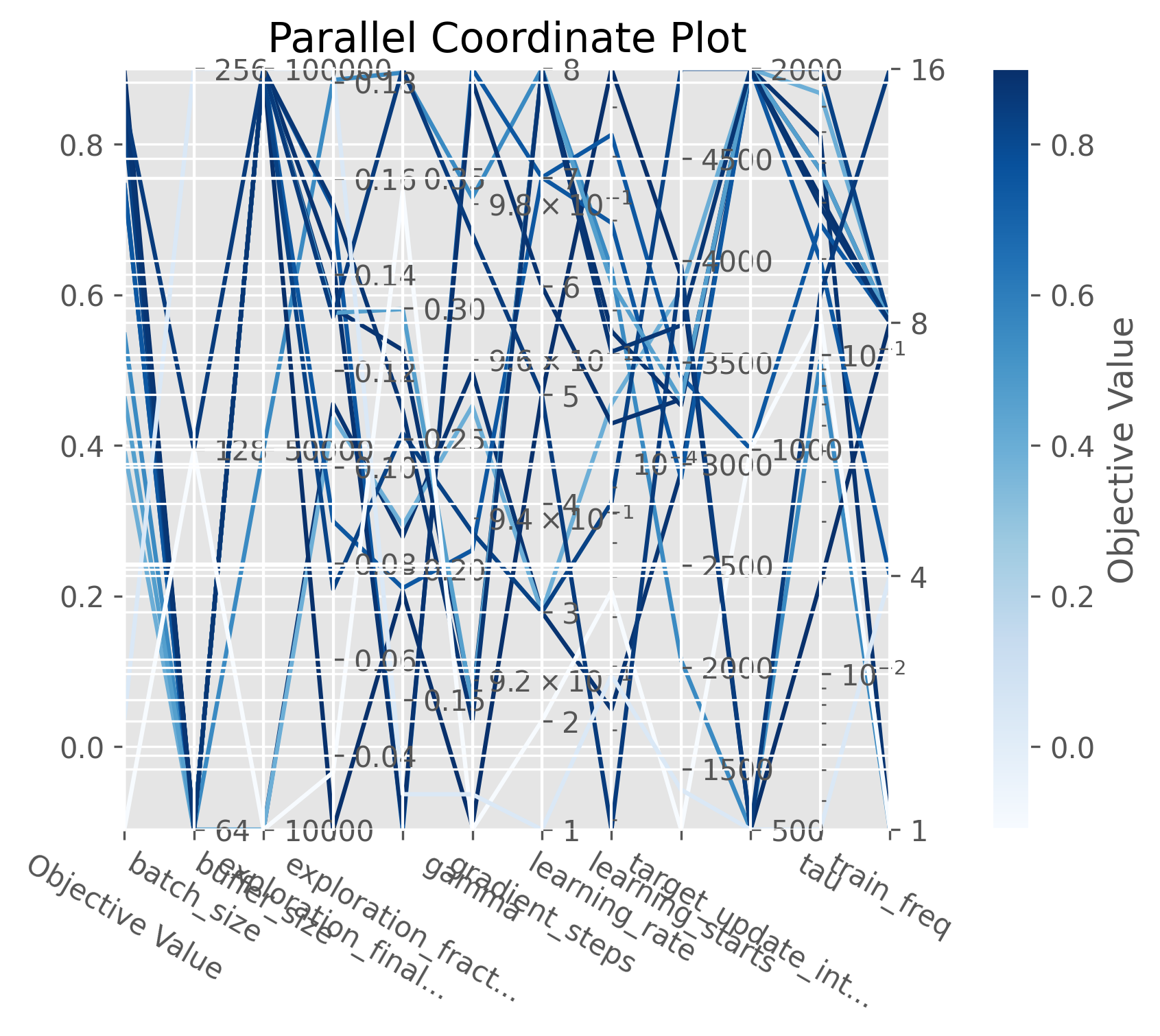}
        \caption{Parallel coordinates of all 25 trials. High-reward lines cluster
        consistently at buffer $= 100{,}000$.}
        \label{fig:parallel}
    \end{minipage}
\end{figure}

These results confirm that the DQN generator operated under a carefully validated
configuration, and that the benchmark results in Section~4 reflect learned policy
behavior rather than favorable default settings.

\subsection{Randomization Test of the DQN Generator}
\label{app:randomization}

To verify that the DQN generator does not exhibit mode collapse, the tendency of a
trained policy to repeatedly produce structurally identical trajectories, a
randomization test was performed over 20 independent episodes on the
\texttt{ToMStoryEnv} environment \cite{towers2024gymnasium}. Three dimensions were
evaluated: action-space coverage, story uniqueness, and character diversity.

All 15 discrete DSL actions were used across 300 timesteps (100\% coverage), with
no single action dominating the distribution. Notably, high-complexity operators such
as \texttt{double\_bluff} (19 times) and \texttt{one\_way\_mirror\_observation}
(24 times) were selected at rates comparable to simpler physical actions, confirming
that the surrogate reward effectively encouraged cognitively complex behavior.
Story lengths ranged from 5 to 16 events (mean: 9.65, $\sigma = 2.26$) and all 20
generated stories were structurally unique with zero duplicates.

\begin{figure}[h]
    \centering
    \begin{minipage}{0.48\textwidth}
        \centering
        \includegraphics[width=\linewidth]{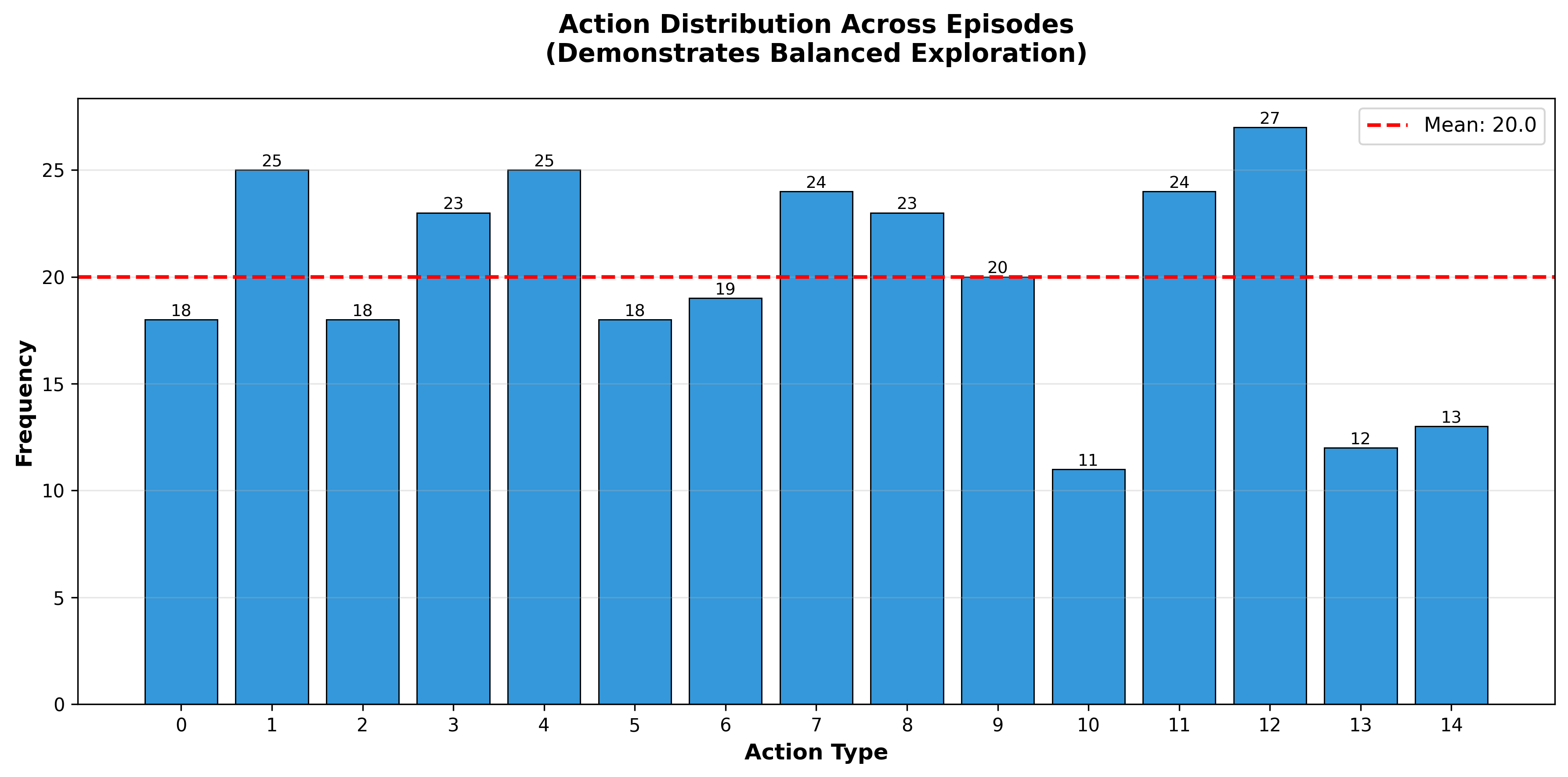}
        \caption{Action distribution across 300 timesteps. All 15 DSL operators
        are represented, confirming full action-space coverage.}
        \label{fig:action_dist}
    \end{minipage}\hfill
    \begin{minipage}{0.48\textwidth}
        \centering
        \includegraphics[width=\linewidth,height=0.75\linewidth,keepaspectratio]{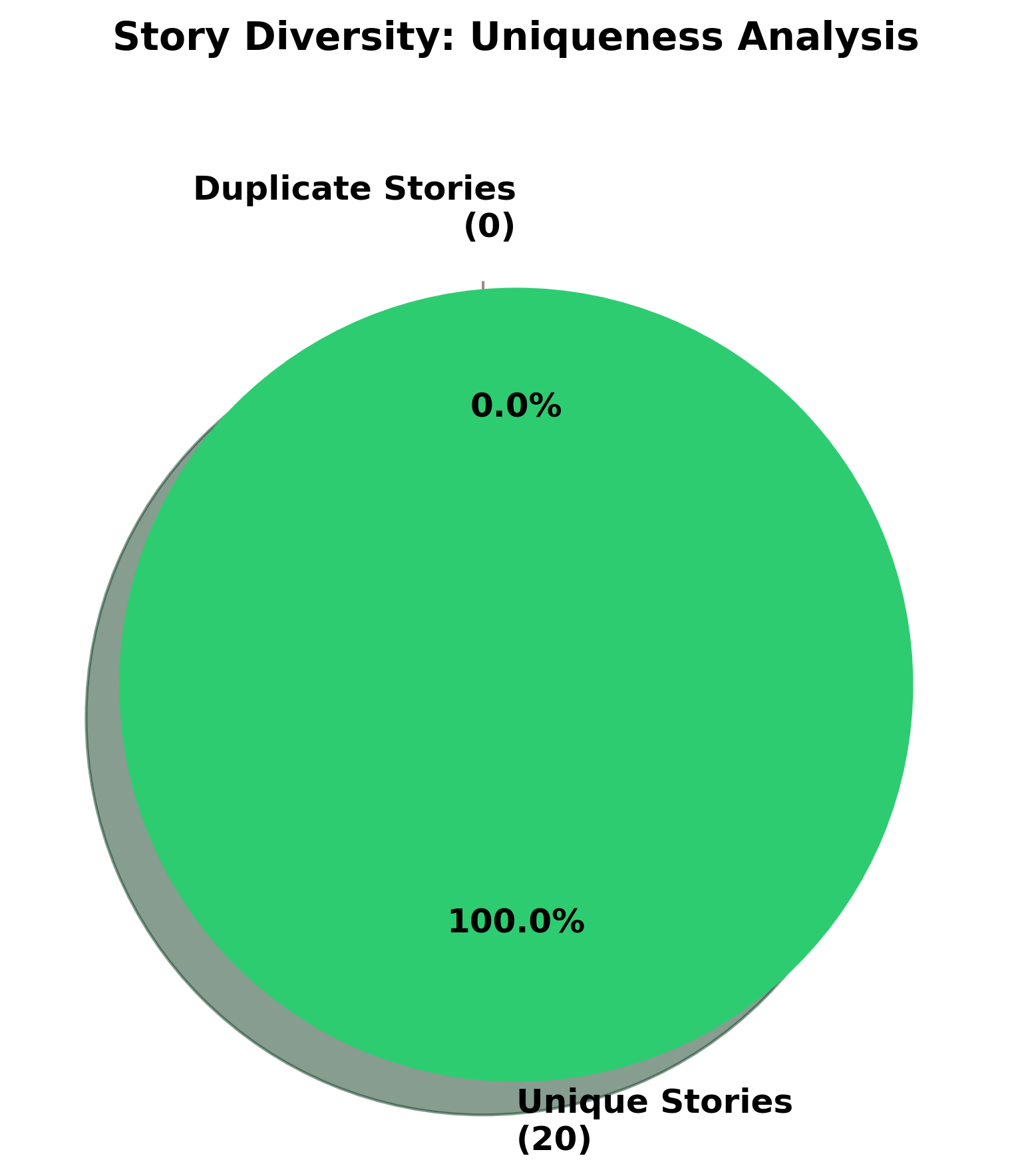}
        \caption{Story diversity metrics, confirming 100\% story uniqueness and
        consistent structural variation across episodes.}
        \label{fig:story_div}
    \end{minipage}
\end{figure}

Character diversity reached 89.7\% (52 unique characters from 58 sampled), with no
single character appearing in more than 3 of the 20 episodes. The randomization test
returned a formal verdict of \textbf{PASS}, confirming that the OSCT dataset was
produced by a generative policy, a necessary precondition for the curriculum
fine-tuning in Section~3.6 to yield a model capable of generalizing across the full
range of Observer-Self Conflict configurations.

\end{document}